\documentclass[journal]{IEEEtran}
\pdfoutput=1
\usepackage{float}
\usepackage{graphicx}
\usepackage{subcaption}
\usepackage{textcomp}
\usepackage{xcolor}
\usepackage{float}
\usepackage{multirow}
\usepackage{csquotes}
\usepackage{url}

\usepackage{tabularx}
\usepackage{booktabs}
\newcolumntype{Y}{>{\raggedleft\let\newline\\\arraybackslash\hspace{0pt}}X}
\newcolumntype{Z}{>{\centering\let\newline\\\arraybackslash\hspace{0pt}}X}

\usepackage{enumitem}
\setlist[itemize]{noitemsep,topsep=0pt}

\newcommand{\ts}{\textsuperscript}

\ifCLASSINFOpdf

\else

\fi

\begin{document}

\title{Iris Liveness Detection using a Cascade of Dedicated Deep Learning Networks.}

\author{Juan Tapia,~\IEEEmembership{Member,~IEEE,}
        Sebastian Gonzalez,~\IEEEmembership{Member,~IEEE,}
        and~Christoph Busch,~\IEEEmembership{Member,~IEEE,}\\ 
   **The following paper is a pre-print. The publication is currently under review for IEEE-TIFS.**     
\thanks{Juan Tapia and Christoph Busch, da/sec-Biometrics and Internet Security Research Group, Hochschule Darmstadt, Germany, e-mail: (juan.tapia-farias@h-da.de, christoph.busch@h-da.de).}
\thanks{Sebastian Gonzalez, R+D Center, TOC Biometrics Company, Santiago, Chile, email: sebastian.gonzalez@toc.cl}
\thanks{Manuscript received xxx; revised xx.}}

\markboth{Journal of \LaTeX\ Class Files,~Vol.~14, No.~8, August~2015}%
{Shell \MakeLowercase{\textit{et al.}}: Bare Demo of IEEEtran.cls for IEEE Journals}

\maketitle

\begin{abstract}
Iris pattern recognition has significantly improved the biometric authentication field due to its high stability and uniqueness. Such physical characteristics have played an essential role in security and other related areas. However, presentation attacks, also known as spoofing techniques, can bypass biometric authentication systems using artefacts such as printed images, artificial eyes, textured contact lenses, etc. Many liveness detection methods that improve the security of these systems have been proposed. The first International Iris Liveness Detection competition, where the effectiveness of liveness detection methods is evaluated, was first launched in 2013, and its latest iteration was held in 2020. This paper proposes a serial architecture based on a MobileNetV2 modification, trained from scratch to classify bona fide iris images versus presentation attack images. The bona fide class consists of live iris images, whereas the attack presentation instrument classes are comprised of cadaver, printed, and contact lenses images, for a total of four scenarios. All the images were pre-processed and weighted per class to present a fair evaluation. This proposal won the LivDet-Iris 2020 competition using two-class scenarios. Additionally, we present new three-class and four-class scenarios that further improve the competition results. This approach is primarily focused in detecting the bona fide class over improving the detection of presentation attack instruments. For the two, three, and four classes scenarios, an Equal Error Rate (EER) of 4.04\%, 0.33\%, and 4,53\% was obtained respectively. Overall, the best serial model proposed, using three scenarios, reached an ERR of 0.33\% with an Attack Presentation Classification Error Rate (APCER) of 0.0100 and a Bona Fide Classification Error Rate (BPCER) of 0.000. This work outperforms the LivDet-Iris 2020 competition results.
\end{abstract}

\begin{IEEEkeywords}
LiveDet, PAD, Presentation Attack Detection, MobileNet.
\end{IEEEkeywords}

\IEEEpeerreviewmaketitle

\section{Introduction}

\IEEEPARstart{I}ris recognition systems has been shown to be robust over time, affordable, non-invasive, and touchless; these strengths will allow it to grow in the market in the coming years. Iris recognition systems are usually based on near-infrared (NIR) lighting and sensors, and have been shown to be susceptible to Presentation Attack Instruments (PAI)~\cite{OpenSource}, where PAI refers to a biometric characteristic or object used in a presentation attack. Presentation Attack Detection (PAD) refers to the ability of a biometric system to recognize PAIs, that would otherwise fool the system into recognizing an illegitimate user as a genuine one, by means of presenting a synthetic forged version of the original biometric trait to the capture device. The biometric community, including researchers and vendors, have thrown themselves into the challenging task of proposing and developing efficient protection mechanisms against this threat~\cite{Galbally}. PAD methods have been suggested as a solution to this vulnerability. Attacks are not restricted to merely theoretical or academic scenarios anymore, as they are starting to be carried out against real-life operations. One example is the hacking of Samsung Galaxy S8 devices with the iris unlock system, using a regular printer and a contact lens. This case has been reported to the public from hacking groups attempting to get recognition for real criminal cases, including from live biometric demonstrations at conferences\footnote{\url{https://www.forbes.com/sites/ianmorris/2017/05/23/samsung-galaxy-s8-iris-scanner-hacked-in-three-simple-steps/#33f150b2ccba}}.

An ideal PAD technique should be able to detect all of these attacks, along with any new or unknown PAI that may be developed in the future~\cite{Hoffman}.

In order to improve PAD methods, a few competitions and databases have been created, such as the LiveDet-Iris\footnote{\url{https://livdet.org/}}. The goal of the Liveness Detection Competition (LivDet-Iris) is to compare biometric liveness detection methodologies, using a standardized testing protocol and large quantities of attack presentation (spoofed) and bona fide presentation samples. This competition has shown that there are still challenges for the detection of iris presentation attacks, mainly when unknown materials or capture devices are used to generate the attacks~\cite{livedet2020}. The results show that even with latest advances in presentation attacks, printed iris PAIs, as well as patterned contact lenses PAIs, are still difficult for software-based systems to detect according with the quality of the images. In LiveDet2017~\cite{livedet2017}, Printed iris images were easier to be differentiated from live images in comparison to patterned contact lenses, as it was also shown in the previous competitions. Some properties of the samples (images) are unknown during training, making the challenge a difficult task, as the winning algorithm did not recognize from 11\% to 38\% of the attack images, depending on the database.  Therefore, the PAD techniques are still an open challenge in NIR, and it has been even less explored in VIS periocular images and multiple capture devices.

The results from the LiveDet2020~\cite{livedet2020} competition indicate that iris PAD is still far from a fully solved research problem. Large differences in accuracy among baseline algorithms, which were trained with significantly different data, stress the importance of access to large and diversified training datasets, encompassing a large number of PAIs. The winning team’s (our method) also achieved the lowest Bona Fide Classification Error Rate (BPCER) of 0.46\%, out of all nine algorithms in the three categories. This aligns well with the operational goal of PAD algorithms to correctly detect bona fide presentations (i.e., and not to contribute to system’s False Alarm Rate), and capture as many attacks as possible.

One of the main challenges to improve PAD systems is the quantity and quality of the data available. Printed images are easy to reproduce with different kinds of paper. Conversely, post-morten images~\cite{TROKIELEWICZ2020103866}, and PAIs such as contact lenses, cosmetic lenses, plastic lenses, all sourced from different brands, are hard to get. Therefore a subject-disjoint dataset containing different iris patterns is difficult to achieve. Alternatively, these datasets can be synthetically created using Deep Learning techniques.

Some relevant techniques, such as morphing~\cite{seibold} and Generative Adversarial Network (GAN)~\cite{survey_gan}, can create very realistic images of faces. It is believed that such algorithms can develop synthetic iris images (NIR/VIS) of several PAIs in order to create challenging databases that can help improving the robustness of PAD systems~\cite{4dcyclegan}. 

\begin{figure*}[!htb]
\centering
	\includegraphics[scale=0.45]{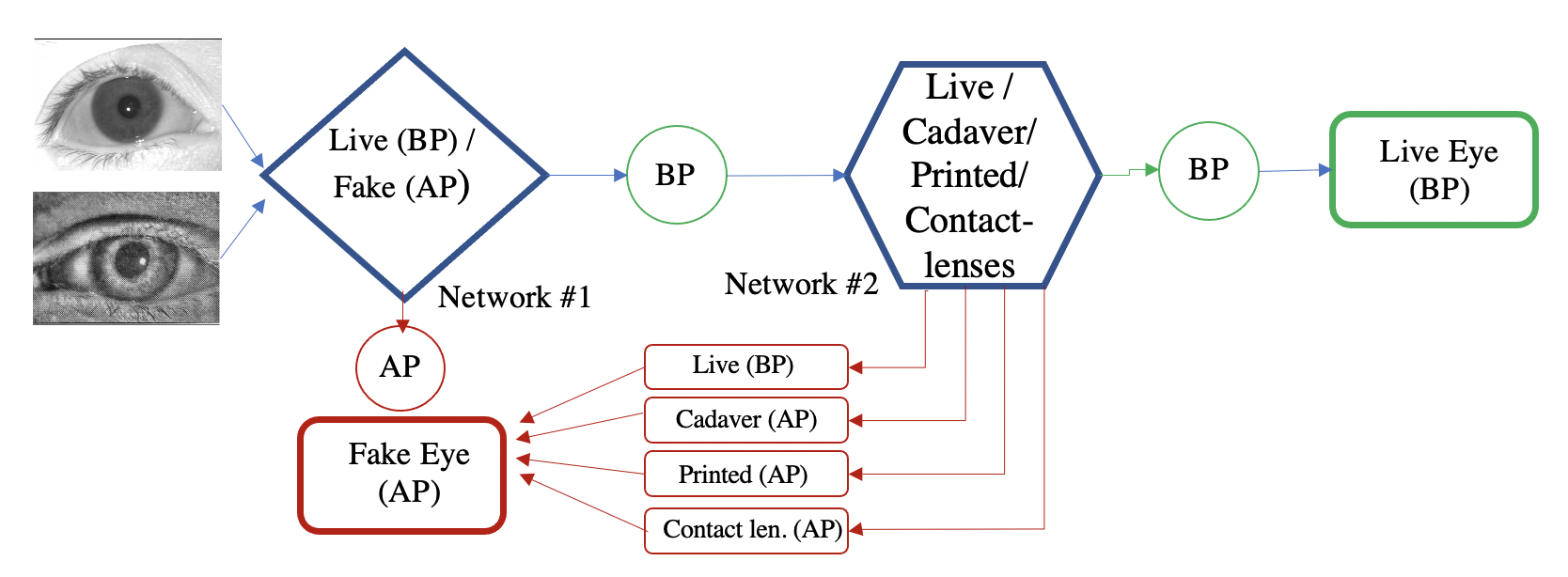}
\caption{Proposed Two-stage serial framework for Presentation Attack Detection. AP: Attack Presentation. BP: Bona fide Presentation.}
\label{fig:real-flow}
\end{figure*}

In this work, a serial, two-stage architecture for classification of bona fide, presentation attack, high-quality printed, and digitally displayed images of LiveDet-2020, plus three complementary databases were explored using deep learning techniques. See Figure \ref{fig:real-flow}. The main contributions of this work can be summarized a follows:

\begin{itemize}
    \item \emph{Architecture}: A serial, two-stage architecture is proposed. This consists of a modified MobileNetV2 model (\enquote{MobileNetv2a}), trained from scratch, which is utilized to differentiate between bona fide presentation and presentation attack. A second MobileNet named \enquote{MobileNetv2b}, trained from scratch for four scenarios, which is then used to detect printed/contact-lenses/cadaver impostor attacks by identifying the physical source of the images. See Figure~\ref{fig:real-flow}.

    \item \emph{Network inputs}: A strong set of experiments of serial and parallel structures of DNNs was evaluated with two, three, and four classes, using NIR images. Bona fide versus contact lenses, print-out, cadavers, electronics and prosthetic displays were used as input to the network. Also, separate and exhaustive experiments were realized using one of these four types of input, and the results were analyzed. 

    \item \emph{Weights}: Balanced class weights were used in order to correctly represent the number of images per class. Most of the spoofing databases are unbalanced according to PA scenarios. Weighted classes help to balance the dataset and to get realistic results.

    \item \emph{Database}: This paper presents two new databases, one database to increase the number of bona fide images (10,000), and a second database to increase the number of printed PAIs with high-quality images (1,800). Both databases will be available to other researchers upon request, for research purposes only (See Section~\ref{sec:database}).

    \item \emph{Data-Augmentation (DA)}: An aggressive DA technique to train the modified MobileNetV2a and MobileNetV2b networks was used. These images allow the network more challenging scenarios considering blurring, Gaussian noise, coarse occlusion, zoom, and others.

    \item \emph{Winning method}: Focused on the correct classification of bona fide images instead of the identification of several PAIs, the serial approach presented in this work reached first place in the LiveDet-2020 competition. Furthermore, the current proposal with three and fourth classes outperforms our results presented in the competition, featuring more challenging scenarios.

    \item \emph{Not self-reported}: The two-stage algorithm presented in this paper was evaluated by the organizers of the competition in an independent test on unknown data; the test data was not available for the participants.
\end{itemize}

\section{Related Work}
\label{sec:relate}

Zou et al.~\cite{4dcyclegan} have presented a novel algorithm, 4DCycle-GAN, for expanding spoofed iris image databases, by synthesizing artificial iris images wearing textured contact lenses. The proposed 4DCycle-GAN follows the Cycle Consistent Adversarial Networks (Cycle-GAN) framework, which translates between one kind of image (bona fide iris images) to another kind (textured contact lenses iris images). Despite the improvements on Conditional Generative Adversial Networks, there are still some open problems that limit its application for image generation. Therefore, the method helps to create and increase the number of images based on conditional GANs while preserving the information in the images of each PAIs in the NIR spectrum. 

Hu et al.~\cite{hu2016iris} investigated the use of regional features in iris PAD (RegionalPAD). Features are extracted from local neighborhoods, based on spatial pyramid (multi-level resolution) and relational measures (convolution on features with variable-size kernels). Several feature extractors, such as Local Binary Patterns (LBP)~\cite{lbp}, Local Phase Quantization (LPQ)~\cite{LPQ}, and intensity correlogram are examined. They used a three-scale LBP-based feature, since it achieves the best performance, as pointed out by the original authors.

Gragnaniello et al.~\cite{gragnaniello2016using} proposes that the sclera region also contains important information about iris liveness (SIDPAD). Hence, the authors extract features from both the iris and sclera regions. The two regions are first segmented, and scale-invariant local descriptors (SID) are applied. A bag-of-feature method is then used to accumulate the features. A linear Support Vector Machine (SVM) is used to perform final prediction. Also, in~\cite{gragnaniello2016biometric}, domain-specific knowledge of iris PAD is incorporated into the design of their model (DACNN). With the domain knowledge, a compact network architecture is obtained, and regularization terms are added to the loss function to enforce high-pass/low-pass behavior. The authors demonstrate that the method can detect both face and iris presentation attacks.

SpoofNets~\cite{kimura20} are based on GoogleNet, and consist of four convolutional layers and one inception module. The inception module is composed by layers of convolutional filters of dimension 1$\times$1, 3$\times$3, and 5$\times$5, executed in parallel. It has the advantage of reducing the complexity and improving the efficiency of the architecture, once the filters of dimension 1$\times$1 help reduce the number of features before executing layers of convolution with filters of higher dimensions.

Boyd et al.~\cite{boyd2020deep} chose the ResNet50 architecture as a backbone to explore whether iris-specific feature extractors perform better than models trained for non-iris tasks. They demonstrated three types of networks: off-the-shelf networks, fine-tuned, and networks trained from scratch, with five different sets of weights for iris recognition. They concluded that fine-tuning an existing network to the specific iris domain performed better than training from scratch.

Yadav et al.~\cite{Yadav_2018_CVPR_Workshops}, a combination of handcrafted and deep-learning-based features was used for iris PAD. They fused multi-level Haralick features with VGG16 features to encode the iris textural patterns. The VGG16 features were extracted from the last fully connected layer, with a size of 4,096, and then reduced to dimensional vector by Principal Component Analysis (PCA).

Nguyen et al.~\cite{PNguyen} proposed a PAD method by combining features extracted from local and global iris regions. First, they trained multiple VGG19~\cite{simonyan2015deep} networks from scratch for different iris regions. Then, the features were separately extracted from the last fully connected layer, before the classification layer of the trained models. The experimental results showed that the PAD performance was improved by fusing the features based on both feature-level and score-level fusion rules.

Kuehlkamp et al.~\cite{andrey} propose an approach for combining two techniques for iris PAD: CNNs and Ensemble Learning. Extensive experimentation was conducted using the most challenging datasets publicly available. The experiments included cross-sensor and cross-dataset evaluations. Results show a varying ability for different BSIF+CNN representations to capture different aspects of the input images. This method outperform the results presented in the LivDet-Iris 2017 competition.

Our approach, presented in the LivDet-Iris 2020 competition, reached the first place with an the Average Classification Error Rate (ACER) of 29.78\%. This method achieved also the lowest Bona Fide Classification Error Rate (BPCER) of 0.46\% out of all nine algorithms in the three categories. This paper show the relevance of focusing mainly in the bona fide images as a \enquote{first-filter}. However, a broad space for improvement was detected in the identification of the PAIs scenarios, specially in cadaver and printed iris images. An Attack Presentation Classification Error Rate (APCER) of 9.87\% was reached for the electronic display PAI, which is lower than all competing algorithms (53.08\% and 83.95\%) by a large margin. 

Based on previous results, this current paper proposes a new framework to improve the PAIs performance per scenarios in order to get a strong PAD method.

The rest of the article is organized as follows: Section~\ref{sec:relate} summarizes the related works on Presentation Attack Detection. The ISO metrics are explained in Section~\ref{sec:metric}. The database description is explained in Section~\ref{sec:database}. The experimental framework is then presented in Section~\ref{sec:metodo}, and the results are discussed in Section~\ref{sec:exp_results}. We conclude the article in Section~\ref{conclusions}.

\section{Databases}
\label{sec:database}

\begin{figure*}[!htb]
\centering
    \subcaptionbox{\centering Live/Real}{\includegraphics[width=0.24\textwidth]{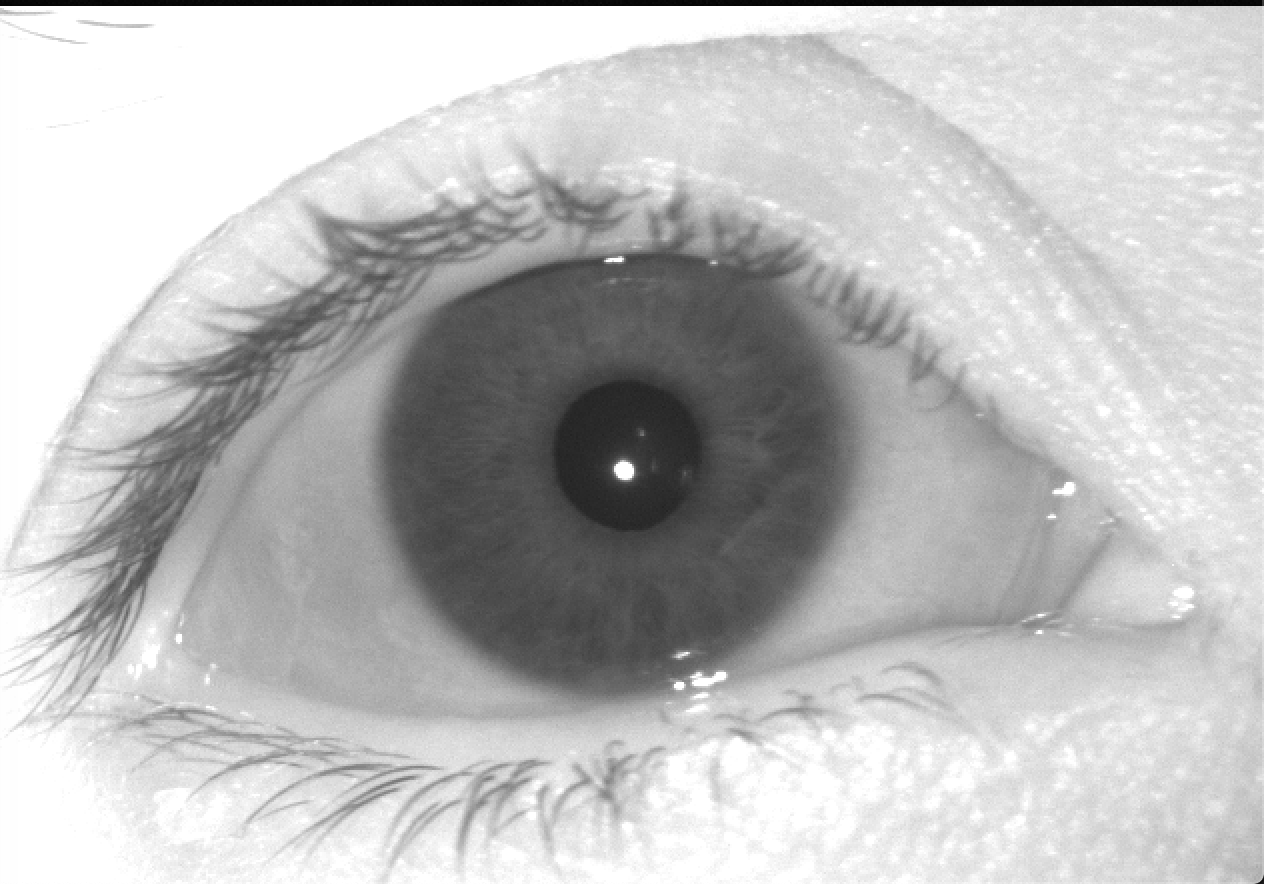}}
    \hfill
    \subcaptionbox{\centering Print-out LivDet-Iris 2020}{\includegraphics[width=0.24\textwidth]{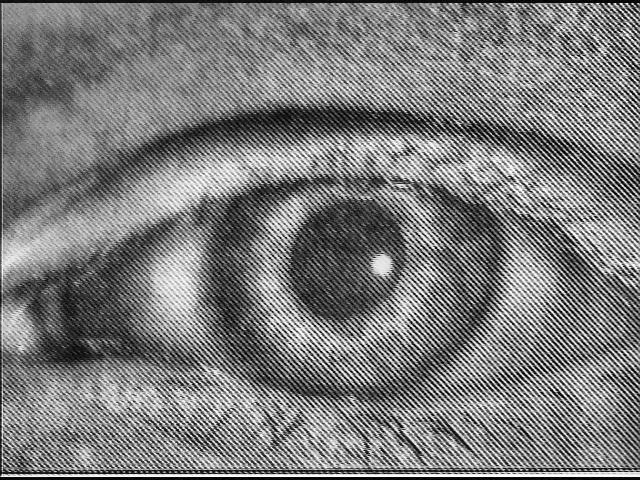}}
    \hfill 
    \subcaptionbox{\centering Print-out Motorola Moto G4 Play}{\includegraphics[width=0.24\textwidth]{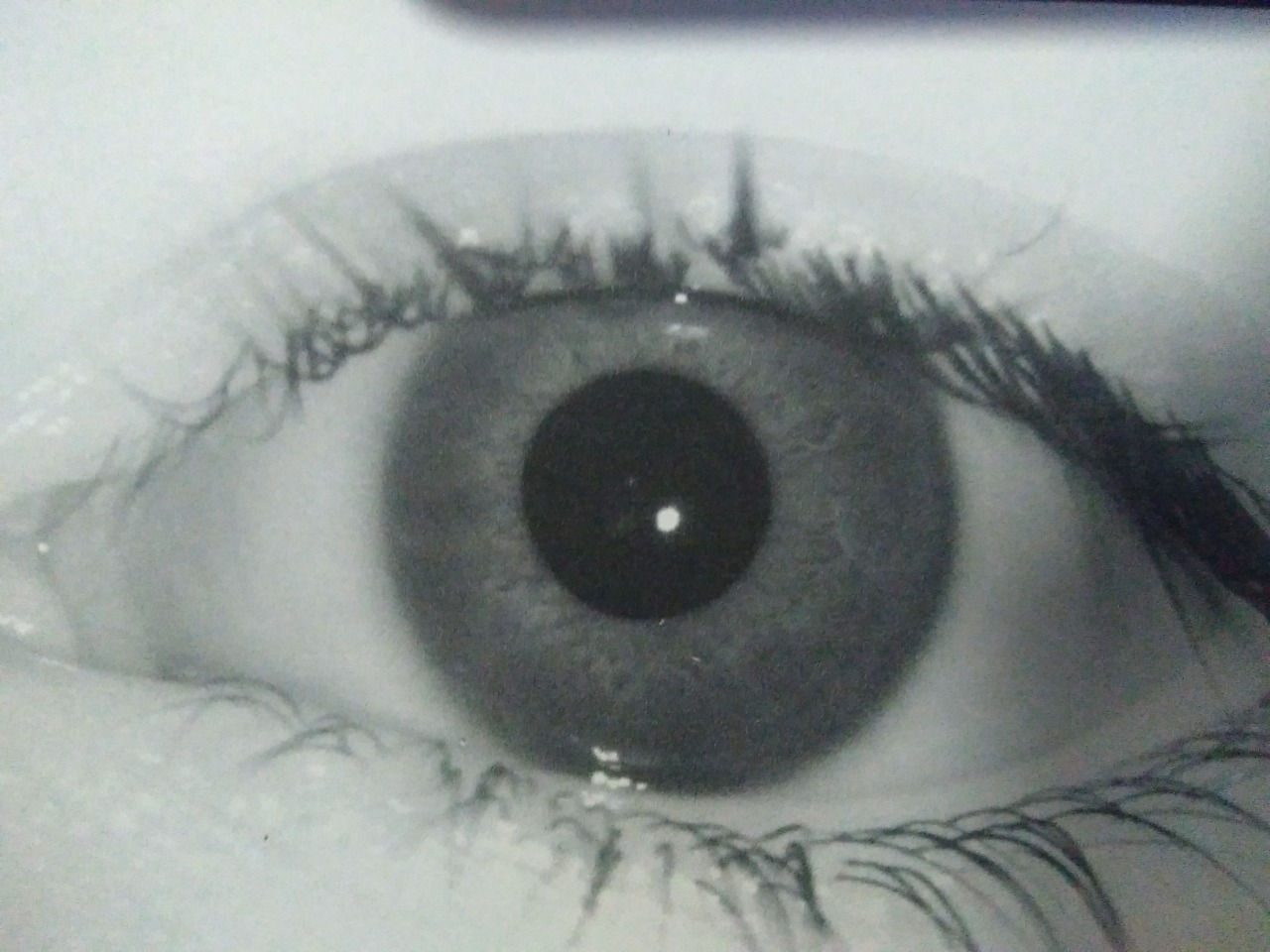}}
    \hfill
    \subcaptionbox{\centering Print-out Nokia 9 PureView}{\includegraphics[width=0.24\textwidth]{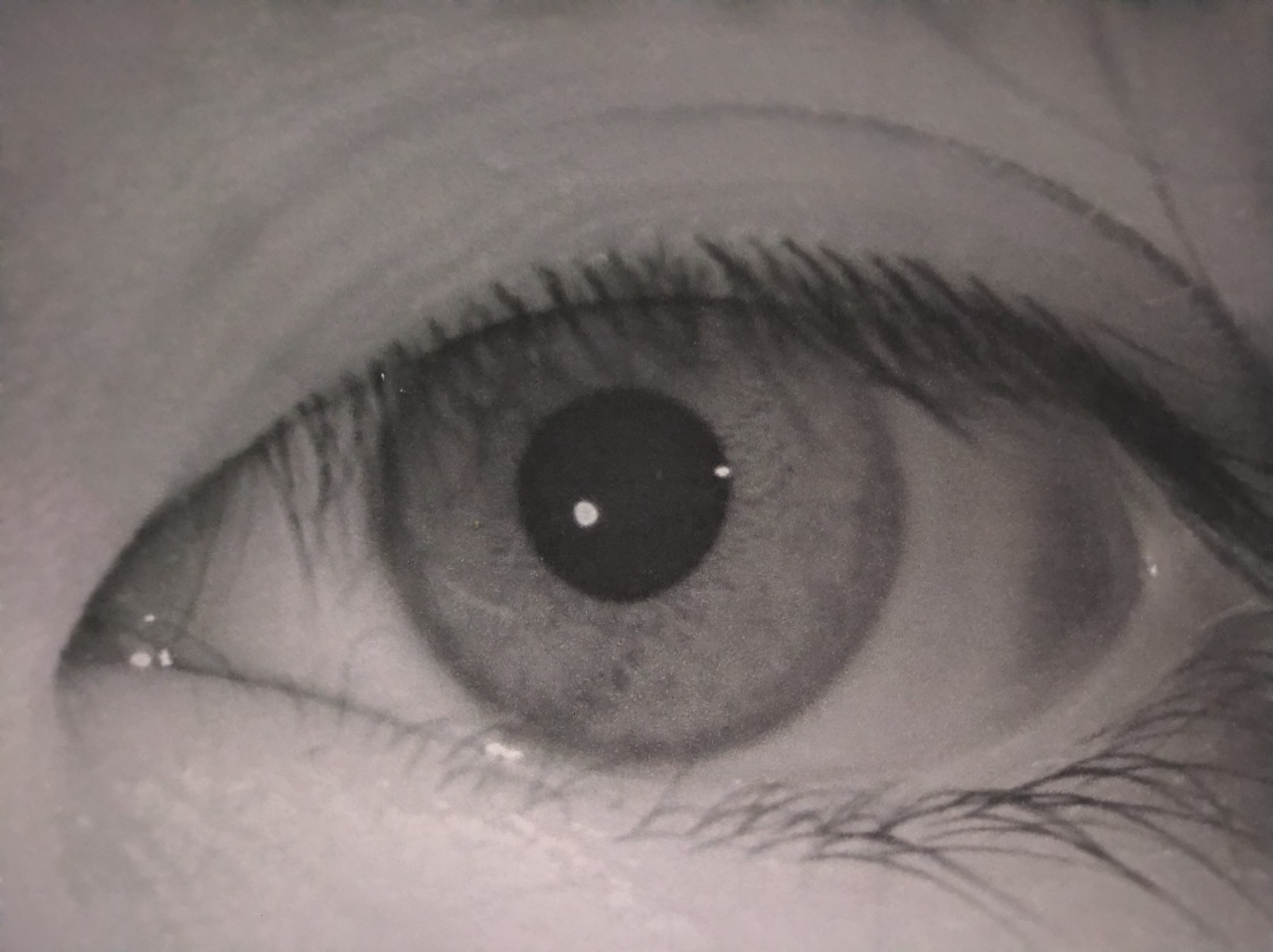}}
    \hfill
    \subcaptionbox{\centering Cadaver (post-mortem subject) eye}{\includegraphics[width=0.24\textwidth]{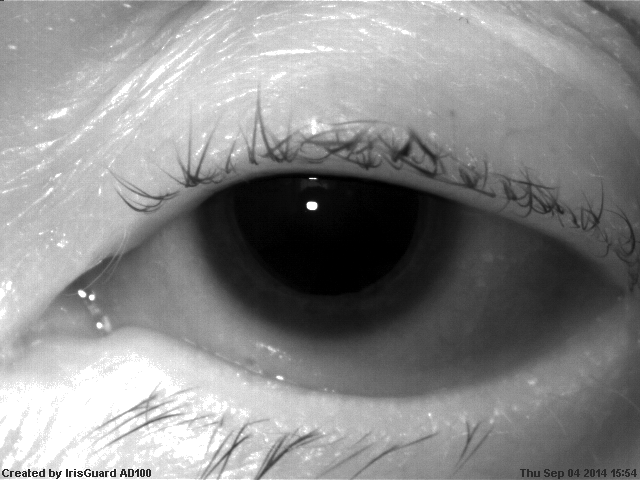}}
    \hfill
    \subcaptionbox{\centering Cosmetic contact lens}{\includegraphics[width=0.24\textwidth]{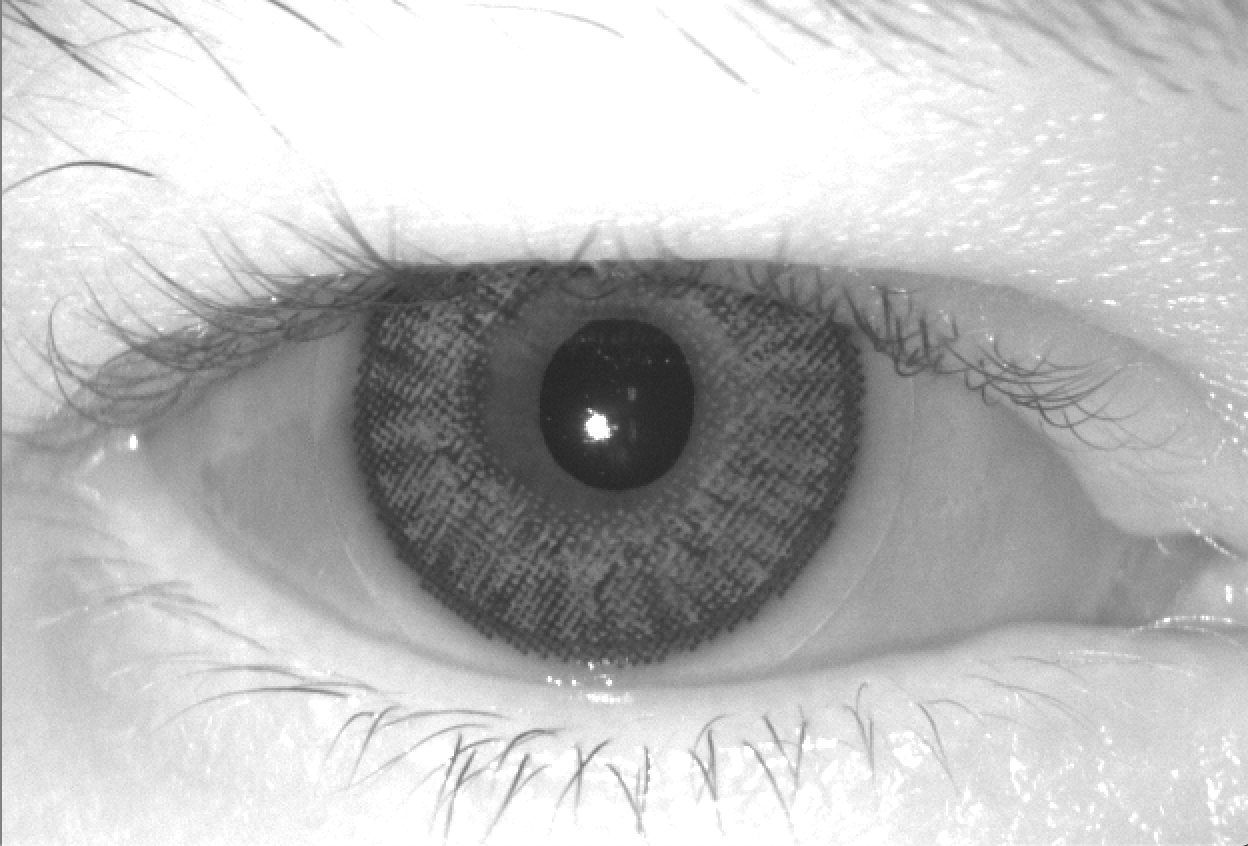}}
    \hfill
    \subcaptionbox{\centering Electronic display}{\includegraphics[width=0.24\textwidth]{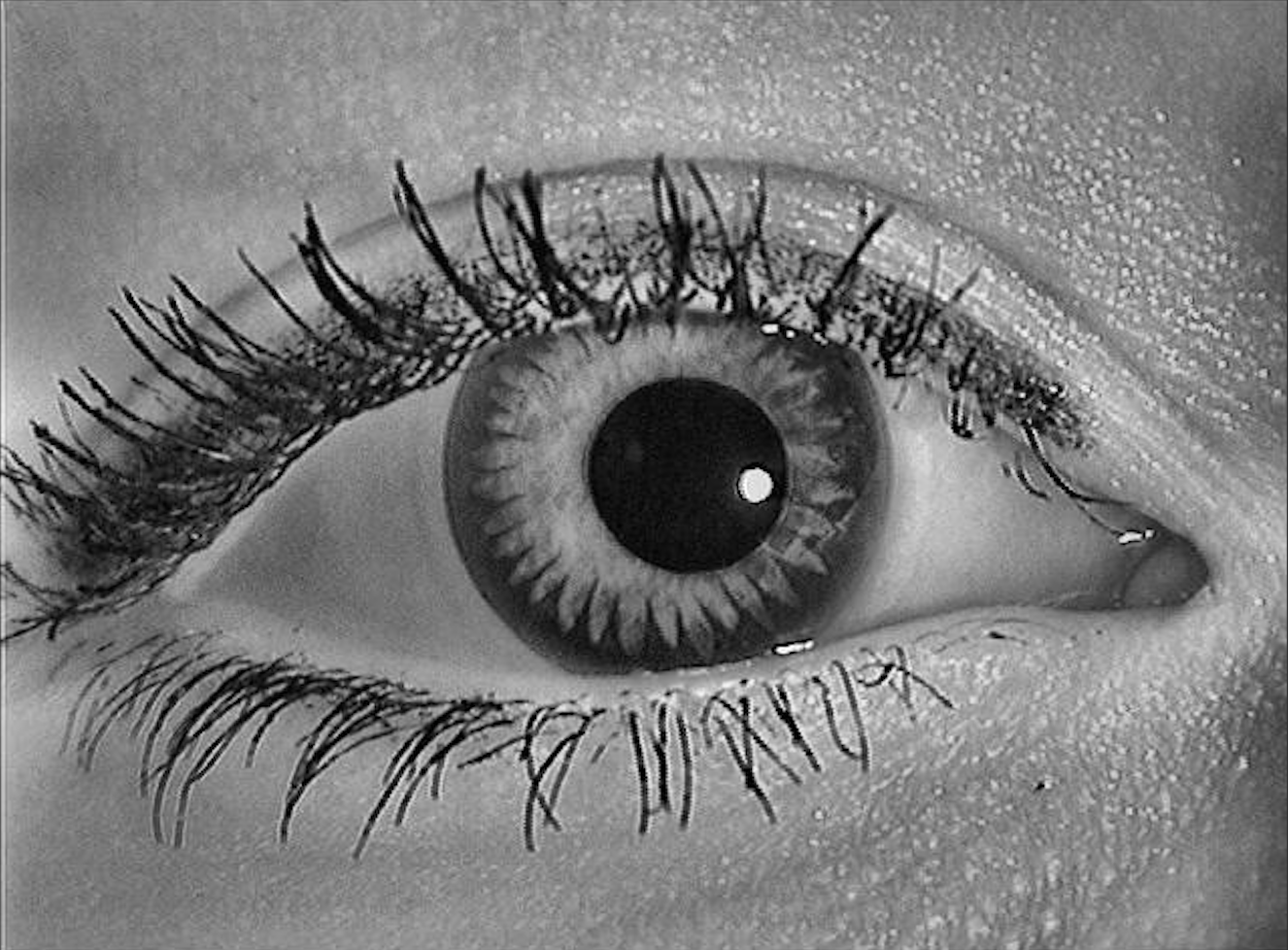}}
    \hfill
    \subcaptionbox{\centering Patterned eye}{\includegraphics[width=0.24\textwidth]{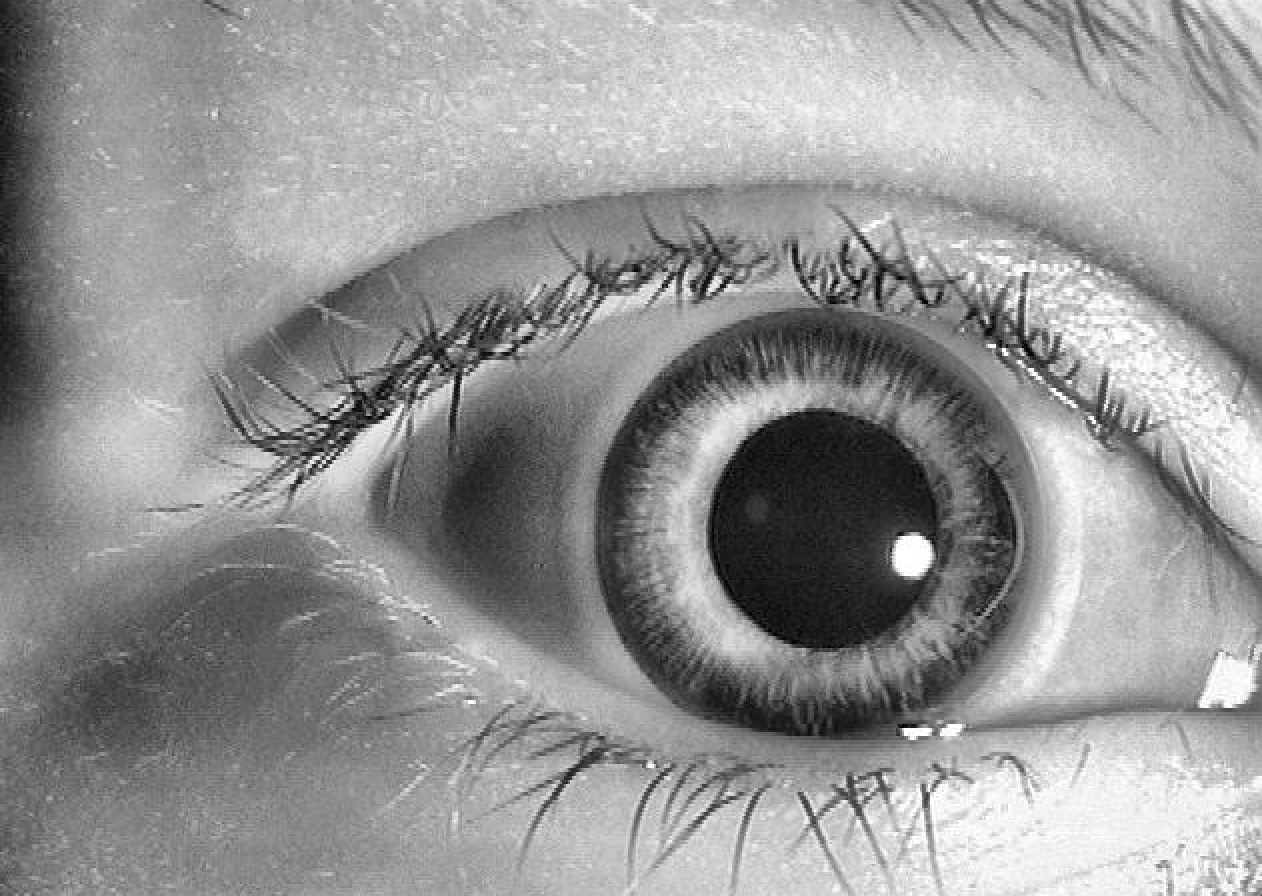}}
\caption{Example images of all presentation attack instruments in the database. Images c, and d show examples of the new PAIs included.}
\label{fig:Example_images}
\end{figure*}

For this work, the LivDet-Iris 2020 competition database was used. In addition, three sets of complementary databases of iris images were also utilized. First, a database of NIR bonafide images, captured using an Iritech TD100 iris sensor with a resolution of 640$\times$480 pixels, called \enquote{Iris-CL1}. A second database, called \enquote{iris-printed-CL1}, containing high-quality presentation attack images of printed PAIs was created. The goal of this database is to increase the challenge of the printed irises scenario, due to the noticeable visible patterns in the printed images from the LivDet-Iris 2020 database, which makes them trivial to distinguish from bona fide images. See Figure~\ref{fig:Example_images}. The iris-printed-CL1 database contains 1,800 images captured with two smartphone devices (900 images each one): a Nokia 9 PureView device, with an image resolution of 1280$\times$957 pixels, and a Motorola Moto G4 Play device, with an image resolution of 1280$\times$960 pixels. Only the red channel was used. These new datasets will be available to others researchers upon request. Figure~\ref{fig:Example_images} present new images of printed scenarios.

The third database is the Warsaw-BioBase-Post-Mortem-Iris v3.0 database~\cite{TROKIELEWICZ2020103866}. This database contains a total of 1,094 NIR images (collected with an IriShield M2120U), and 785 visible-light images (obtained with Olympus TG-3), collected from 42 post-mortem subjects. This database was not fully available for the competition.

The LivDet-Iris 2020 database included five different PAIs, each with a different level of challenge: printed eyes, textured contact lens, electronic display, fake/prosthetic eyes, and printed with add-ons, and a small number of cadaver eyes. The printed image dataset is of a very low resolution. No specific training dataset was prepared for the competition. A total of 11,918 images were made available.

The competition was different from previous editions in regards to the training dataset. the participants were encouraged to use all the data available to them (both publicly available and proprietary) to make their solutions as effective and robust as possible. The entirety of previous LivDet-Iris benchmarks were also made publicly available~\cite{yambay2014schuckers,7947701,livedet2017}. Additionally, the competition organizers shared five examples of each PAI (samples which were not used later in evaluations) to help the competitors familiarize themselves with the test data format (pixel resolution, bits per pixel used to code the intensity, etc.).

Table~\ref{table:Train_Dataset} shows a summary of the all databases available for training in LivDet-Iris 2020. The datasets of presentation attack instruments (PAIs) were specifically created for the development of PAD methods. With the evolving of PAIs, the datasets include new challenges. A detailed technical summary of the available datasets can be found in~\cite{boyd2020deep,czajka2018presentation}.

\begin{table}[!htb]\def\tabularxcolumn#1{m{#1}}
\centering
\caption{Training Dataset Summary -- 11,918 images. Fa, PrD, Pr, represent: Fake/Prosthetic Display and Printed Add-ons. BP: Bona fide presentation. AP: Attack Presentation.}
\label{table:Train_Dataset}
	\begin{tabularx}{\linewidth}{Xllr}
	\toprule
	\multicolumn{1}{c}{\textbf{Dataset}} & \textbf{Class} & \multicolumn{1}{c}{\textbf{PAIs}} & \textbf{Num Images}\\
	\midrule
	LDet-Iris-2013-Clarkson & BP & --- & 516\\
    LDet-Iris-2015-Clarkson & BP & --- & 541\\
    LDet-Iris-2017-Clarkson & BP & --- & 3,954\\
    LDet-Iris-2020-Clarkson & BP & --- & 5\\
    LDet-Iris-2020-Notre Dame & BP & --- & 5\\
    LDet-Iris-2013-Clarkson & AP & Cont. Lens & 840\\
    LDet-Iris-2015-Clarkson & AP & Cont. Lens & 824\\
    LDet-Iris-2017-Clarkson & AP & Cont. Lens & 1,887\\
    LDet-Iris-2020-Notre Dame & AP & Cont. Lens & 5\\
    LDet-Iris-2015-Clarkson & AP & Printouts & 1,077\\
    LDet-Iris-2017-Clarkson & AP & Printouts & 2,254\\
    LDet-Iris-2020-Clarkson & AP & Printouts & 1\\
    LDet-Iris-2020-Clarkson & AP & Elec. Display & 1\\
    LDet-Iris-2020-Clarkson & AP & Fa, PrD, Pr & 3\\
    LDet-Iris-2020-Warsaw & AP & Cadaver Eyes & 5\\
    \midrule
    \multicolumn{3}{c}{\textbf{Total}} & \textbf{11,918}\\
	\bottomrule
	\end{tabularx}
\end{table}

Table~\ref{tab:my-sum2} shows a summary of all datasets available from LivDet-Iris 2020, plus Cadaver images, iris-CL1, and iris-printed-CL2. The new total count of images available is 27,964. This is more than two times the number of images shown in Table~\ref{table:Train_Dataset}.

\begin{table}[!ht]\def\tabularxcolumn#1{m{#1}}
\centering
\caption{Summary of the new, complete database, with 27,964 images divided in train, test, and validation.}
\label{tab:my-sum2}
\setlength{\tabcolsep}{4.6pt}
	\begin{tabularx}{\linewidth}{ccrrrrc}
	\toprule
	\textbf{Class} & \textbf{PAIs} & \multicolumn{1}{c}{\textbf{Train}} & \multicolumn{1}{c}{\textbf{Val}} & \multicolumn{1}{c}{\textbf{Test}} & \multicolumn{1}{c}{\textbf{Num Im.}} & \textbf{Sensors}\\
	\midrule
	BP & --- & 6,694 & 1,062 & 5,773 & 13,530 & \begin{tabular}[c]{@{}c@{}}LG4000\\AD 100\\iCam 700\\TD100\end{tabular}\\
	\midrule
    AP & Cadaver & 448 & 531  & 754 & 1,773 & \begin{tabular}[c]{@{}c@{}}IriTech\\IriShield\end{tabular}\\
    \midrule
    AP & \begin{tabular}[c]{@{}c@{}}Cont. Lenses\\Textured\end{tabular} & 3,583 & 900 & 3,244 & 7,727 & \begin{tabular}[c]{@{}c@{}}LG4000\\AD 100\\iCam 700\\TD100\\MotoG4\\Gplay\end{tabular}\\
    \midrule
    AP & \begin{tabular}[c]{@{}c@{}}Printed\\Prosthetic\\Display\end{tabular} & 4,090 & 1,896 & 2,305 & 8,291 &  \begin{tabular}[c]{@{}c@{}}Iris ID\\iCAM700\end{tabular}\\
    \midrule
    \multicolumn{2}{c}{\textbf{Total}}& \textbf{11,810} & \textbf{4,384} & \textbf{11,770} & \textbf{27,964} & \\
	\bottomrule
	\end{tabularx}
\end{table}

\subsection{Unknown Scenarios}

An unknown state-of-the-art dataset of spoofed images was used to measure the performance of our proposal~\cite{bsif_bowyer}. This dataset includes 900 images of textured contact lenses produced by Cooper and J\&J (i.e. not represented in the training set of LivDet-Iris 2020) and 900 images of authentic irises. It has been shown that PAD methods do not generalize well to a brand of textured contact lenses not seen in the training data, thus the unknown dataset of spoofed images is used in the context of cross-dataset testing~\cite{bsif_bowyer}.

\subsection{Data Augmentation}

\begin{figure*}[!htb]
\centering
	\includegraphics[scale=0.51]{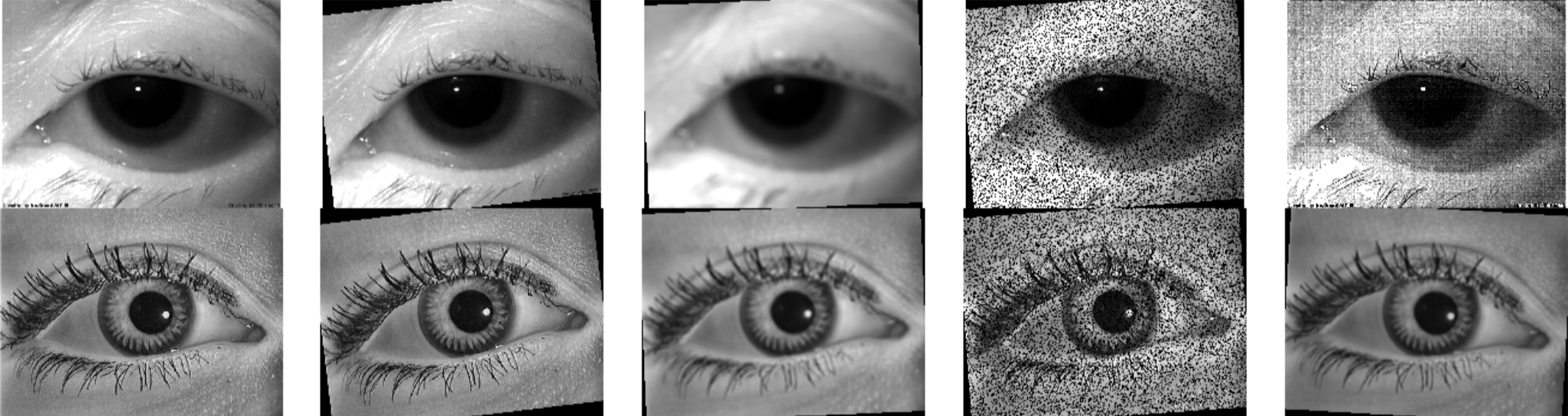}
\caption{Examples of the aggressive data augmentation applied randomly to live and fake images. Left: original images, rotation, blurring, Gaussian noise filter, and Filter Edge Enhance.}
\label{fig:DA}
\end{figure*}

An aggressive data augmentation (DA) method was applied when training the modified MobileNetV2 networks. All the images were normalized using a histogram equalization algorithm, and then weighted. A large number of images, with several operations such as affine transformations, perspective transformations, contrast changes, Gaussian noise, random dropout of image regions, hue/saturation changes, cropping/padding, and blurring were included in the train dataset. These DA operations are based on the imgaug library~\cite{imgaug}, which is optimized for high performance. This improves the quality of the training results by using very challenging images. Examples of some augmented images are shown in Figure~\ref{fig:DA}.

\section{Metrics}
\label{sec:metric}

The ISO\_/IEC 30107-3 standard \footnote{\url{https://www.iso.org/standard/67381.html}} presents methodologies for the evaluation of the performance of PAD algorithms for biometric systems. The APCER metric measures the proportion of attack presentations---for each different PAI---incorrectly classified as bona fide (genuine) presentations. This metric is calculated for each PAI, where ultimately the worst-case scenario is considered. Equation~\ref{eq:apcer} details how to compute the APCER metric, in which the value of $N_{PAIS}$ corresponds to the number of attack presentation images, where $RES_{i}$ for the $i$th image is $1$ if the algorithm classifies it as an attack presentation (spoofed image), or $0$ if it is classified as a bona fide presentation (real image)~\cite{marcel2019handbook}.

\begin{equation}\label{eq:apcer}
    APCER=\frac{1}{N_{PAIS}}\sum_{i=1}^{N_{PAIS}}(1-RES_{i})
\end{equation}

Additionally, the BPCER metric measures the proportion of bona fide (live images) presentations mistakenly classified as attacks presentations to the biometric algorithm, or the ratio between false rejection to total genuine attempts. The BPCER metric is formulated according to equation~\ref{eq:bpcer}, where $N_{BF}$ corresponds to the number of bona fide (live) presentation images, and $RES_{i}$ takes identical values of those of the APCER metric.

\begin{equation}\label{eq:bpcer}
    BPCER=\frac{\sum_{i=1}^{N_{BF}}RES_{i}}{N_{BF}}
\end{equation}

These metrics effectively measure to what degree the algorithm confuses presentations of spoofed images with real images, and vice versa. Furthermore, the Average Classification Error Rate (ACER) is also used. This is computed by averaging the APCER and BPCER metrics, as shown in equation~\ref{eq:acer}. This evaluates the overall system performance.

\begin{equation}\label{eq:acer}
    ACER=\frac{APCER+BPCER}{2}
\end{equation}

Note that ACER has been deprecated in ISO/IEC 30.107-3. Only for the purpose of comparing with the state of the art, the ACER was computed.

A Detection Error Trade-off (DET) curve is also reported for all the experiments. In the DET curve, the Equal Error Rate (EER) value represents the trade-off when the APCER is equal to the BPCER. Values in this curve are presented as percentages.

\section{Methodology}
\label{sec:metodo}

In this section, we introduce our baseline by utilizing a fine-tuned network, and a new network trained from scratch. Then, the detailed description of the used convolutional layers is presented.

\subsection{Networks}

MobileNet~\cite{howard2017mobilenets} is based on a streamlined architecture to build lightweight deep neural networks. This allows for usage in environments with limited resources, such as mobile applications, while achieving with state-of-the-art performance for tasks such as classification. A modified MobileNet was used to detect live and spoofed images. 

For this work, ImageNet~\cite{deng2009imagenet} weights were initially used for transfer learning. However, the results of fine-tuning the network would worsen proportionally to the amount of layers that were frozen. Therefore training the networks from scratch resulted in a better classification performance overall. This is explained in more detail in Section~\ref{sec:exp_results}

In addition, two different network architectures are used in this work: MobileNetV2a and a modified MobileNetV2b. MobileNetV2a was trained from scratch, based on bona fide and fake scenarios only, whereas MobileNetV2b was introduced based on live, patterned contact lenses, printed, and cadaver scenarios.

\subsection{Image Pre-processing}

All the images in the database were pre-processed using an adaptive histogram algorithm to improve the gray-scale intensity. Later, a weighted factor per each class was applied. Also, a higher number of filters was applied, using MobileNetV2 alpha parameter, from the standard 1.0 up to 1.4. Both methods are leveraged to create a two-stage classifier that can detect bona fide and spoofing scenarios. All the images were resized to 224$\times$224 and 448$\times$448 according to the experiments.

\subsection{Contrast Limited Adaptive Histogram Equalization (CLAHE)}

In order to improve the quality of the images and highlight texture-related features, the CLAHE algorithm was applied. This algorithm divides an input image into an M$\times$N grid. Afterwards, it applies equalization to each cell in the grid, enhancing global contrast and definition in the output image. All the images were divided in 8$\times$8 sized cells.

\subsection{Class Weights}

A weight factor was estimated for each class according to the number of images of the class, helping in balancing the database. Thus, each class is correctly represented in the error estimation. See Equation~\ref{eq1}.

\begin{equation}
\label{eq1}
    Weight_i = \frac{Nsamples}{Nclasses \times samples_i}
\end{equation}

Where $Weight_i$ is the weight for class $i$, $Nsamples$ is the total number of images in the database, $Nclasses$ is the total number of classes in the database, and $samples_i$ is the number of samples of class class $i$. The weight values associated to each classes are the following: 

\begin{itemize}
    \item Class 0, Cadaver: 4.4162
    \item Class 1, Live: 0.5787
    \item Class 2, Pattern: 1.0133
    \item Class 3, Printed: 0.9443
\end{itemize}

\subsection{Alpha Values}

The number of parameters and number of multiply-adds can be modified by using the alpha parameter, which increases/decreases the number of filters in each layer of the MobileNet network. By altering the image size and alpha parameter. This alpha value is known as the width multiplier in the original MobileNet implementation:

\begin{itemize}
    \item If $alpha = 1.0$, the default number of filters from the original MobileNet paper are used at each layer.
    \item If $alpha < 1.0$, proportionally decreases the number of filters in each layer.
    \item If $alpha > 1.0$, proportionally increases the number of filters in each layer.
\end{itemize}

\section{Experiment and Results}
\label{sec:exp_results}

The approach presented in this work takes into account the variability of the PA images, and the number of images per class. These images present a problem for the classifier because the PAIs are not equally represented (for instance only five images of cadaver eyes were available for LivDet-Iris 2020). Considering this imbalance, our strategy is primarily focused on classifying bona fide images with high precision first, and attack presentation images second. Therefore, our first approach was training a network with only two classes. Then, a second network was trained from scratch with three and four classes, increasing the number of filters (alpha 1.4) and weighting each class according to the numbers of images per scenario. To study these limitations and improve performance for these aforementioned scenarios, five experiments were developed in order to analyze the best hyper-parameter configuration of MobileNetV2. A combination of serial and parallel DNNs was used, trained from scratch. A grid search was used to determine the learning rate, number of epochs, global pooling operation, alpha value, and input size of images. All the experiments employ the CLAHE algorithm and the class weight balancing operation. All the networks were trained with a limit of 200 epochs, using an early stopping method in case the measured performance would stop improving. The image input sizes utilized were 224$\times$224 and 448$\times$448 pixels. All the experiments used the same number of images.

\subsection{Experiment 1}

A traditional MobileNetv2 network was used, trained with fine-tunning techniques. Several tests were performed, sequentially freezing an additional MobileNetV2 block in each one, from the bottom of the network to the top. For this experiment the images were grouped in two classes: Live and Fake. The Fake dataset considers the images all PAIs: Contact Lenses (CL), Printout (Pr), Electronic displays (EDs), Prostetic Display (PD) and Cadaver Eyes (CE). 

\subsection{Experiment 2}

A modified MobileNetv2a network was trained from scratch. For this experiment, the images were again grouped in two classes: BP (Bona fide presentations) and AP (Attack presentation with various PAI). The AP dataset is comprised of all PAI classes: Contact Lenses (CL), Printout (Pr), Electronic displays (EDs), Prostetic Display (PD) and Cadaver Eyes (CE).

\subsection{Experiment 3}

For this experiment, a modified MobileNetv2b network was trained from scratch. The images were grouped in three classes this time: Bona fide, Contact lenses (patterned) and Printouts. 

\subsection{Experiment 4}

A modified MobileNetv2b network were trained from scratch. The images in this experiment were grouped into four classes: Bona fide, Contact lenses, Printouts, and Cadaver. 

\subsection{Experiment 5}

This experiment evaluated the feasibility of our two-stage proposed method, trained with two classes, to detect unknown attacks presentations over traditional PAI species. In particular, we tackle the challenging scenario where the PAI species remain unknown, and is not part of the PAD algorithm training set. For this experiment, we used the unknown test dataset available from the UND database~\cite{bsif_bowyer}.

\subsection{Results}

In this section, we report the best results for each experiment. Adam optimization performed better than SGD and RMSprop. The best initial learning rate was $1 \times 10^{-5}$. Global max pooling performs better than global average pooling. An alpha value of 1.0 performed better with two class scenarios, with an input image size of $224 \times 224$, whereas an alpha value of 1.4 with an input image size of $448 \times 448$ performed better for three and four class scenarios.

Table~\ref{tab:summary} shows an overview of the results for two class scenarios trained with fine-tuning, and two, three, and four class scenarios trained from scratch. Rows 3 to 7 show the fine-tuning results. Only the results for layers 10, 19, and 28 are included, due to the degradation in performance that was proportional to the amount of bottom layers from the network that were frozen. We infer this is probably because the pre-trained ImageNet~\cite{deng2009imagenet} weights were not trained using images of spoof NIR eyes, or anything similar. Overall, for fine-tuning, the best results were obtained when freezing only the first MobileNetV2 block (T2212), using Adam optimization, resulting in an APCER of 4.29\%, and a BPCER of 3.69\%.

\begin{table}[!htb]\def\tabularxcolumn#1{m{#1}}
\centering
\caption{Summary of the results for two, three, and four classes. In bold are highlighted the best results. POOL: global pooling operation used. FT: fine tuning training; number of blocks frozen.}
\label{tab:summary}
\setlength{\tabcolsep}{5.6pt}
	\begin{tabularx}{\linewidth}{crrrlll}
	\toprule
	\textbf{Model} & \begin{tabular}[r]{@{}r@{}}\textbf{ACER}\\(\%)\end{tabular} & \begin{tabular}[r]{@{}r@{}}\textbf{APCER}\\(\%)\end{tabular} & \begin{tabular}[r]{@{}r@{}}\textbf{BPCER}\\(\%)\end{tabular} & \multicolumn{1}{c}{\textbf{POOL}} & \multicolumn{1}{c}{\textbf{OPT}} & \multicolumn{1}{c}{\textbf{FT}}\\
	\midrule
	\multicolumn{7}{c}{\textbf{2 CLASSES}}\\
	\midrule
    \textbf{T2212} & \emph{3.99} & \emph{4.29} & \emph{3.69} & \emph{AVG} & \emph{ADAM} & \emph{1\ts{st} only}\\
    T0710 &  9.50 & 12.44 &  6.57 & AVG & SGD  & 1\ts{st} \& 2\ts{nd}\\
    T1619 &  6.51 &  7.42 &  5.60 & AVG & ADAM & 1\ts{st} \& 2\ts{nd}\\
    T0123 & 10.30 & 11.59 &  5.01 & AVG & SGD  & 1\ts{st} to 3\ts{rd}\\
    T1022 &  6.33 & 11.05 & 16.10 & AVG & ADAM & 1\ts{st} to 3\ts{rd}\\
    T1905 &  5.44 &  6.49 &  4.38 & AVG & SGD  & NONE\\
    T1310 &  6.55 &  5.30 &  7.81 & AVG & SGD  & NONE\\
    \textbf{T0730} & \emph{2.81} & \emph{3.92} & \emph{1.70} & \emph{AVG} & \emph{ADAM} & \emph{NONE}\\
    \midrule
    \multicolumn{7}{c}{\textbf{3 CLASSES}}\\
    \midrule
    \textbf{T2057} & \emph{0.50} & \emph{1.00} & \emph{0.00} & \emph{AVG} & \emph{ADAM} & \emph{NONE}\\
    T2056 & 1.70 & 0.65 & 2.47 & AVG & ADAM & NONE\\
    T1713 & 2.30 & 1.84 & 2.75 & AVG & ADAM & NONE\\
    T1202 & 2.31 & 0.87 & 3.76 & AVG & ADAM & NONE\\
    \midrule
    \multicolumn{7}{c}{\textbf{4 CLASSES}}\\
    \midrule
    T0729 & 7.80 & 14.55 & 1.23 & AVG & ADAM & NONE\\
    T2202 & 7.53 & 13.35 & 1.71 & MAX & ADAM & NONE\\
    \textbf{T1721} & \emph{4.92} & \emph{8.26} & \emph{1.58} & \emph{MAX} & \emph{ADAM} & \emph{NONE}\\
    T1628 & 5.38 & 8.35 & 2.41 & MAX & SGD & NONE\\
	\bottomrule
	\end{tabularx}
\end{table}

\begin{figure*}[!htb]
    \hfill
    \begin{minipage}{.5\linewidth}
    \centering
	    \includegraphics[scale=0.16]{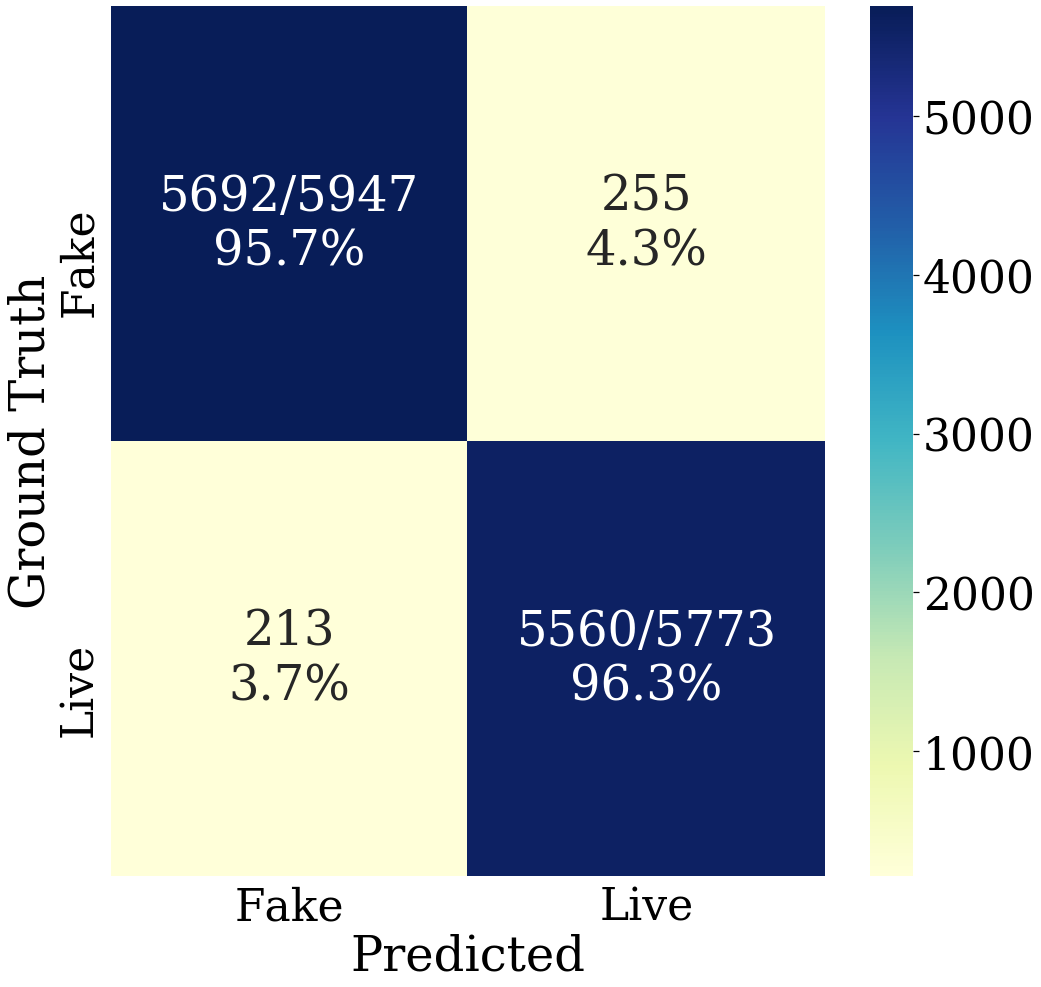}
    \end{minipage}%
    \hfill
    \begin{minipage}{.5\linewidth}
    \centering
        \includegraphics[scale=0.26]{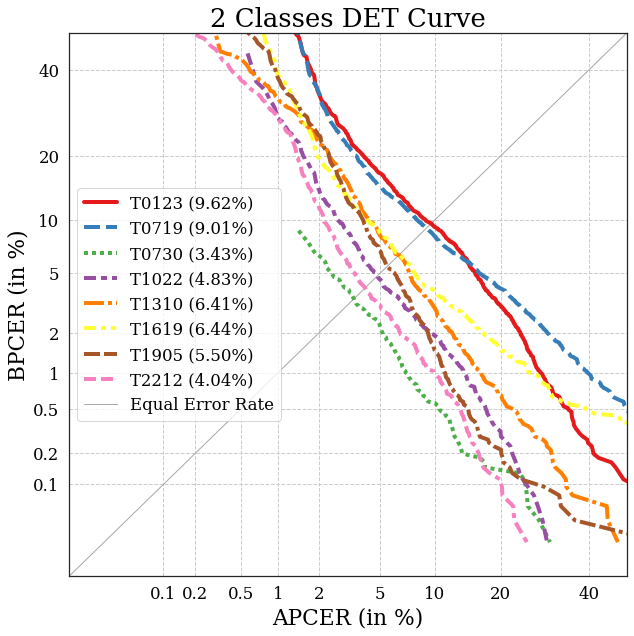}
    \end{minipage}%
    \hfill
\caption{Results of the PAD method with two classes for model T0730. Left: confusion matrix for two class test, PA (fake) and bona fide (live). Right: DET curve for the best results. The number in parenthesis corresponds to the EER in percentage. The lowest EER reached was 3.43\%.}
\label{fig:mc2}
\end{figure*}

Figure~\ref{fig:mc2} shows the best result for two class scenarios, trained from scratch. This allows us to focus on identifying BP images versus AP (fake) images. In this figure, a confusion matrix considering these two classes is shown. Additionally, a Detection Error Trade-off (DET) curve is presented. Several approaches were tested, where the best result reaches an EER of only 3.43\% (green curve), using an alpha value of 1.4, an initial learning rate of $1\times 10^{-5}$, and the Adam optimization algorithm.

\begin{figure*}[!htb]\centering

\centering
	\includegraphics[scale=0.16]{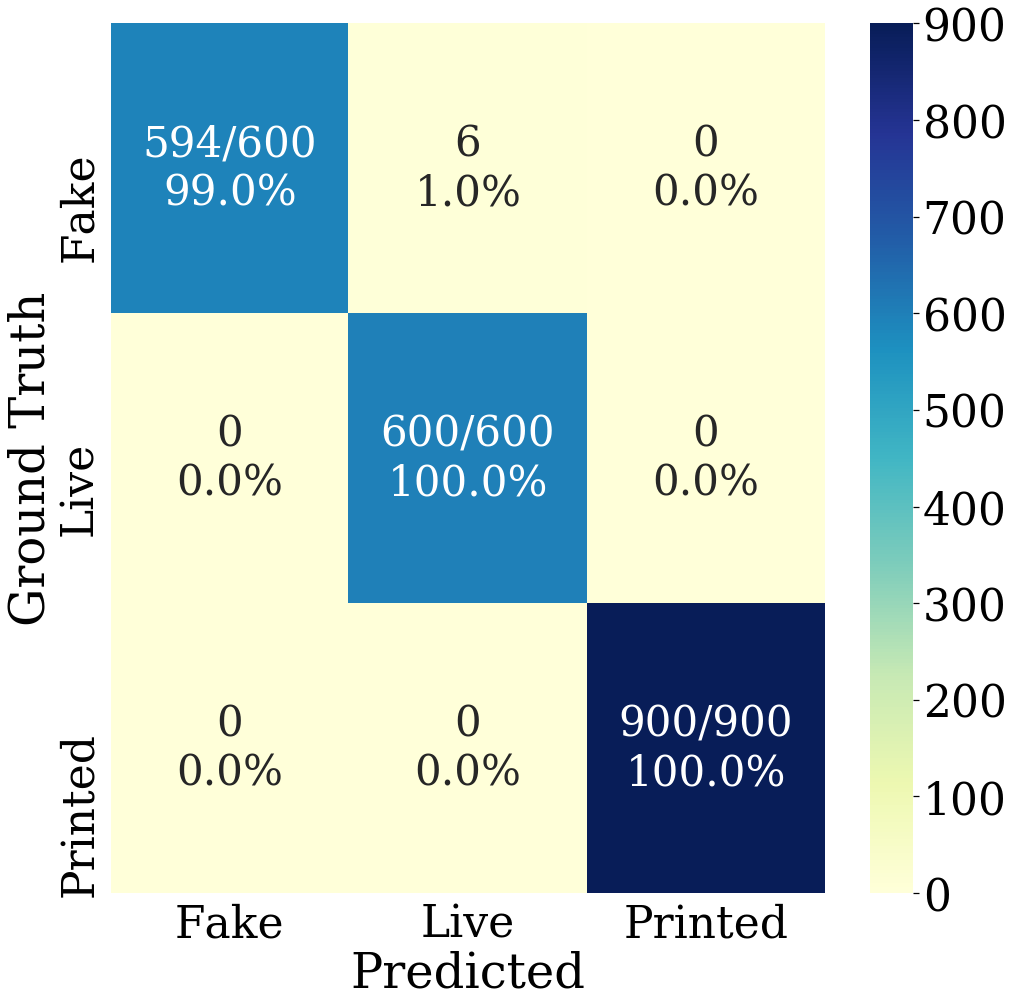}%
	\includegraphics[scale=0.16]{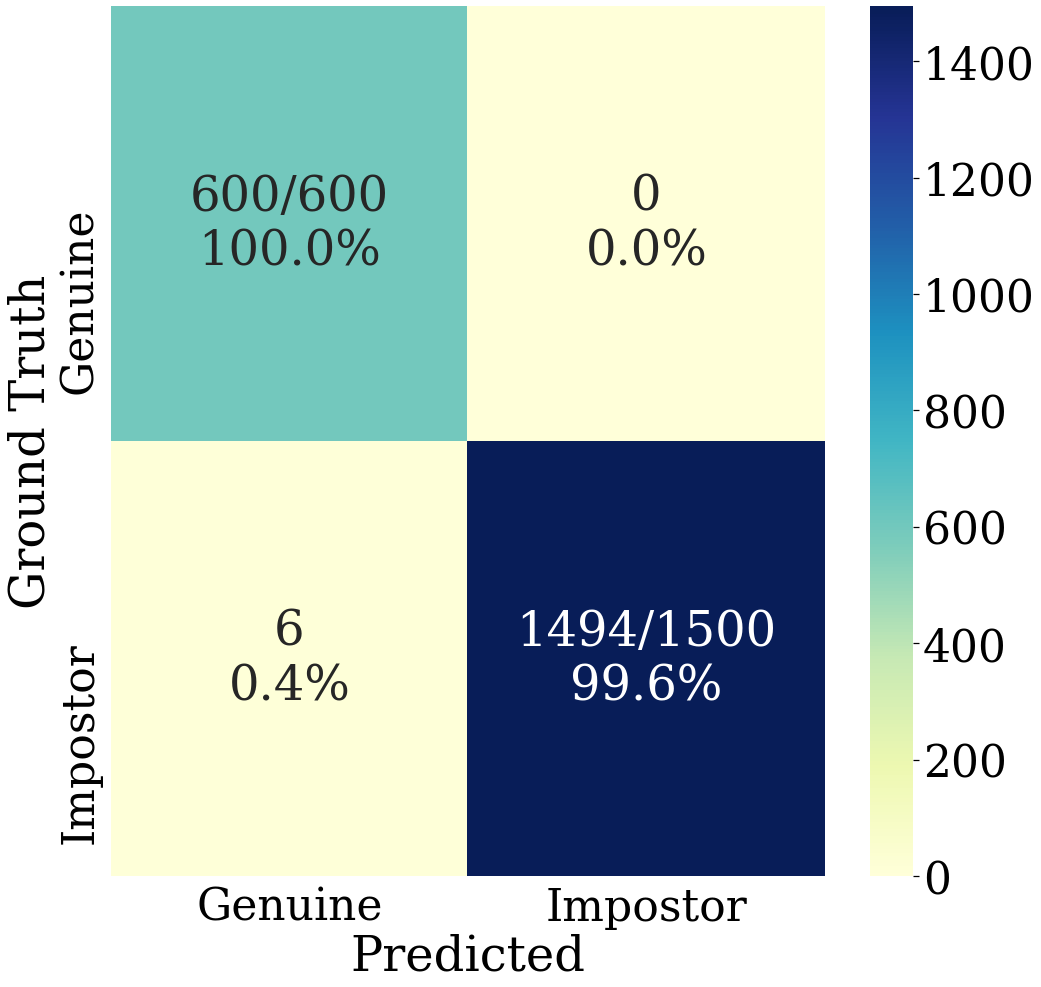}%
	\includegraphics[scale=0.26]{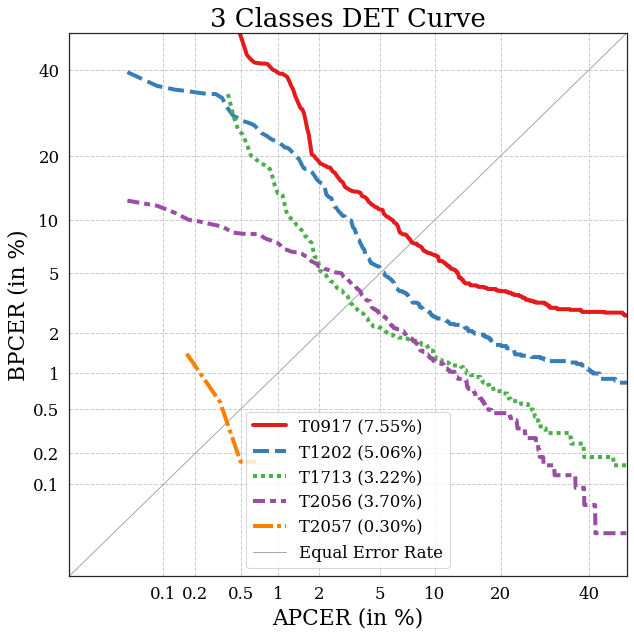}%
\caption{Results of the PAD method with three classes for model T2057. Left: confusion matrix considering each PAI independently. The second confusion matrix considers the bona fide class versus the fusion of all PAIs. The number in parenthesis corresponds to the EER in percentage. The lowest EER reached was 0.30\%.}
\label{fig:mc3a}
\end{figure*}

Figure~\ref{fig:mc3a} shows the best result for three class scenarios: PA, BP, and Contact Lenses. In this figure, a confusion matrix considering these three classes is shown. Furthermore, a confusion matrix showing genuine (BP), and impostor classes is presented. In this case, the impostor class encompasses both PA and Contact Lenses PAIs. Additionally, a Detection Error Trade-off (DET) curve is also shown. The best result reaches an EER of only 0.30\% (orange curve), using an alpha value of 1.4, an initial learning rate of $1\times 10^{-5}$, and the Adam optimization algorithm.

\begin{figure*}[!htb]
\centering
	\includegraphics[scale=0.17]{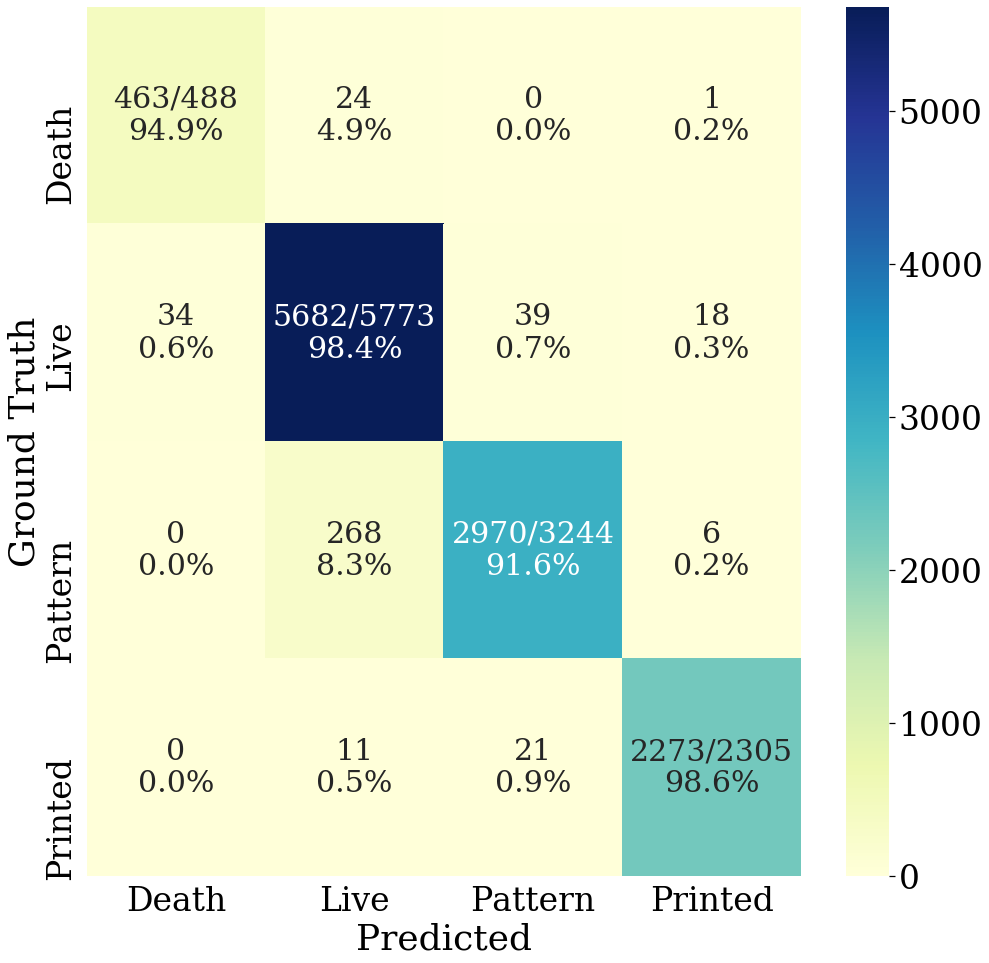}%
	\includegraphics[scale=0.17]{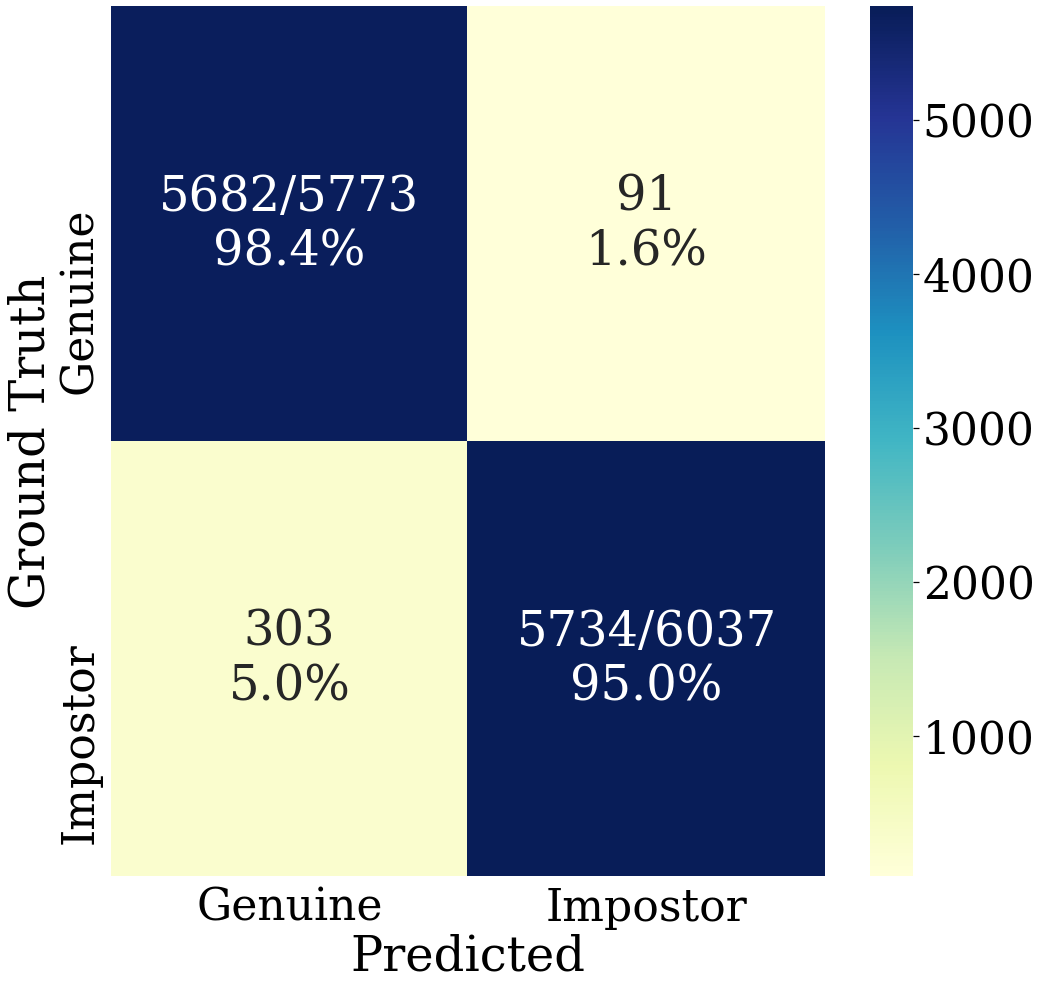}%
	\includegraphics[scale=0.27]{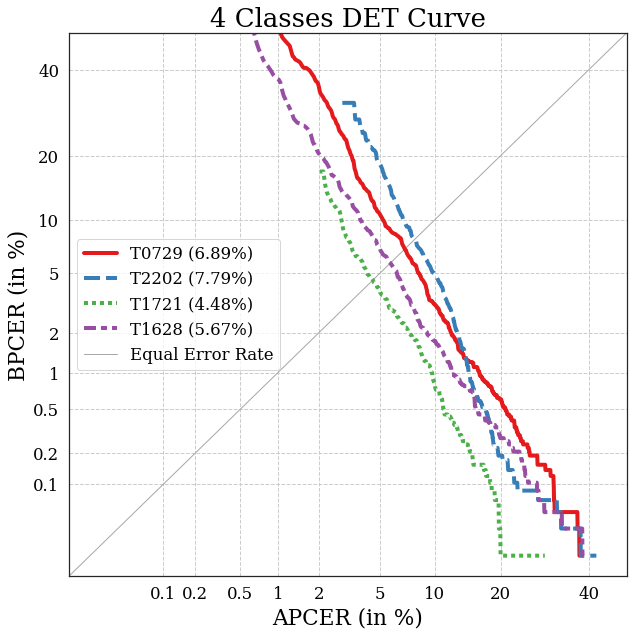}%
\caption{Results of the PAD method with four classes for model T1721. Left: confusion matrix considering each PAI independently. The second confusion matrix considers the bona fide class versus the fusion of all PAIs. The number in parenthesis corresponds to the EER in percentage. The lowest EER reached was 4.48\%.}
\label{fig:mc4a}
\end{figure*}

Figure~\ref{fig:mc4a} shows the best result for four class scenarios: BP, Printed, Contact lenses, and Cadaver. Likewise, two confusion matrices, one showing four classes, and the other grouping all PAIs under the \enquote{impostor} class, are presented. A Detection Error Trade-off (DET) curve is also shown. The best result for this experiment reaches an EER of only 4.48\% (green curve), using an alpha value of 1.4, an initial learning rate of $1\times 10^{-5}$, and the Adam optimization algorithm.

\begin{figure}[!htb]
\centering
	\includegraphics[scale=0.26]{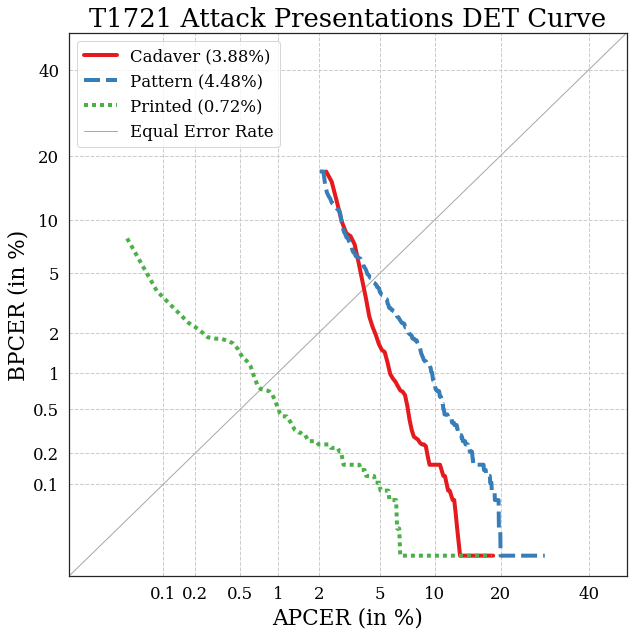}
\caption{DET curve for each PAI for the four classes model.}
\label{fig:attack6}
\end{figure}

Figure~\ref{fig:attack6} shows the performance for each PAI for four class scenarios: BP, Printed, Contact lenses, and Cadaver. The best result corresponds to the printed PAI, with a EER of only 0.72\% (green curve), whereas the worse PAI performance was for patterned contact lenses (ERR of 4.48\%) followed by the cadaver PAI (ERR of 3.88\%).

\begin{figure}[!htb]
\centering
	\includegraphics[scale=0.28]{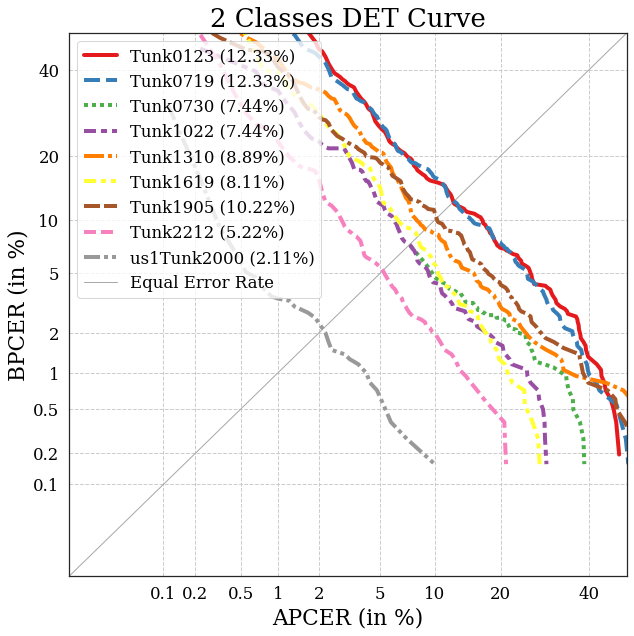}
\caption{DET curves for unknown species.}
\label{fig:attack_unknown}
\end{figure}

Figure~\ref{fig:attack_unknown} shows the performance for unknown species form the UND Database~\cite{bsif_bowyer}. Our proposal was evaluated with a challenging pattern contact lenses scenario. To that end, eight experiments, according to experiment 1, were evaluated. The leave-one-out protocol was applied, using all of the models of experiment 1, and evaluated with experiment 5. The Equal Error Rates reached are in the range of 2.11\% to 12.33\%. The best results show the robustness of our method to unknown scenarios.

Table~\ref{tab:camparison} shows a comparison with the state-of-the-art methods. Our proposal improve the results of the LivDet-Iris 2020 Competition.

\begin{table}[!htb]\def\tabularxcolumn#1{m{#1}}
\centering
\caption{Comparison with the state of the art. Results are shown in \%.}
\label{tab:camparison}
\setlength{\tabcolsep}{4.2pt}
	\begin{tabularx}{\linewidth}{rrrr}
	\toprule
	\textbf{Method} & \textbf{APCER (\%)} & \textbf{BPCER (\%)} & \textbf{ACER (\%)}\\
	\midrule
    USACH/TOC     & 59.10 & 0.46  & 29.78\\
    FraunhoferIGD & 48.68 & 11.59 & 30.14\\
    Competitor 3  & 57.8  & 40.31 & 49.06\\
    NP PAD        & 57.21 & 0.71  & 28.96\\
    MSU PAD Alg1  & 4.67  & 0.56  & 2.61 \\
    MSU PAD Alg 2 & 2.76  & 1.61  & 2.18 \\
    DACNN         & 55.2  & 16.39 & 35.8 \\
    SIDPAD        & 49.85 & 39.96 & 44.9 \\
    Regional PAD  & 62.42 & 23.80 & 43.11\\
    \midrule
    \textbf{Our method \#1 -- 2 Classes} & 3.03 & 1.70 & 2.81\\
    \textbf{Our method \#2 -- 3 Classes} & \emph{1.00} & \emph{0.00} & \emph{0.50}\\
    \textbf{Our method \#3 -- 4 Classes} & 8.35 & 2.41 & 5.38\\
	\bottomrule
	\end{tabularx}
\end{table}

\begin{figure}[!htb]
\centering
	\includegraphics[scale=0.32]{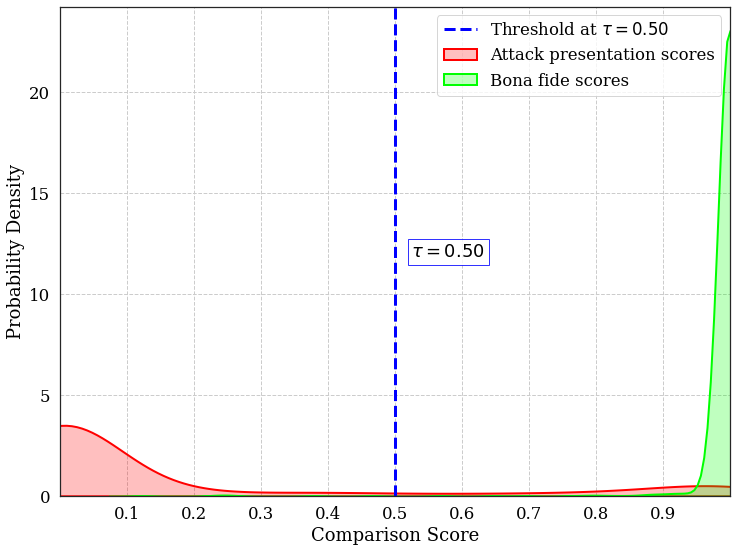}\\
	\includegraphics[scale=0.32]{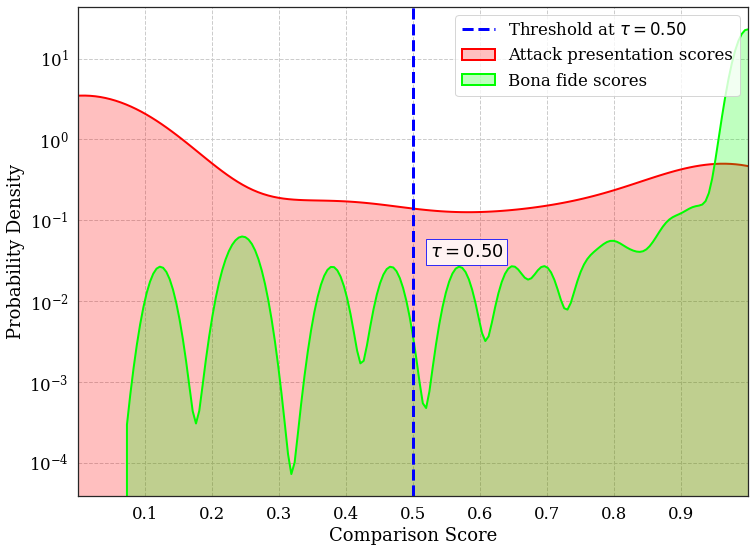}%
\caption{KDE distribution of Attack Presentation Scores versus Bona fide scores. Top: linear scale. Bottom: logarithmic scale. The threshold value (shown in blue) is defined as 0.5 by default. This operating point can be adjusted depending on requirements and use case, subject to the trade-off between APCER and BPCER.}
\label{fig:kde}
\end{figure}

Figure~\ref{fig:kde} shows two Kernel Density Estimation (KDE) plots, in linear and logarithmic scale for the ordinate respectively (the abscissa is shown in linear scale for both), for the best two-classes model (T2057) according to Figure~\ref{fig:mc2}, with a EER of 0.30\%.

\section{Conclusion}
\label{conclusions}

Existing studies in the iris presentation attack detection literature are based on the assumption that the system encounters a specific iris presentation attack. However, this may not be the case in real-world scenarios, where the iris recognition system may have to handle multiple kinds of presentation attacks, including unknown scenarios. To address this challenge, we propose a framework focused on detecting live images, which means optimizing the models for a lower BPCER score. For this approach, we developed the largest iris presentation attack database by combining several other databases. This database are also available to other researchers by requests.

Our suggested networks, when trained from scratch, allow us to improve the results of the LivDet-Iris 2020 competition by using more challenging PAIs. When using fine tuning, model performance worsens in proportion to the amount of layers from the network that were frozen. Nonetheless, results using fine tuning are competitive with the literature.

According to our results, an image input size of $224\times224$ is enough for the successful classification of bona fide images. However, for presentation attack instruments, the results were improved when using an image input size of $448\times448$. This shows that the extra detail from higher resolution images contain relevant features for PAIs classification.

Overall the best result reached was with three scenarios, obtaining an APCER 1.00\%, a BPCER of 0.00\%, and an ACER of 0.50\%. These results were obtained using a threshold value of 0.5; for other use cases---for example, biometric verification systems in banking---a more strict operating point could be used, obtaining a lower APCER but higher BPCER. The best model proposed with three classes reached an EER of 0.33\%. This work outperforms the LivDet-Iris 2020 competition results, serving as the latest evaluation of iris PAD on a large spectrum of presentation attack instruments.

\section*{Acknowledgment}
This research work has been partially funded by the German Federal Ministry of Education and Research and the Hessian Ministry of Higher Education, Research, Science and the Arts within their joint support of the National Research Center for Applied Cybersecurity ATHENE and TOC Biometrics company.

\bibliographystyle{IEEEtran}
\bibliography{references.bib}

\begin{IEEEbiography}[{\includegraphics[width=1in,height=1.25in,clip,keepaspectratio]{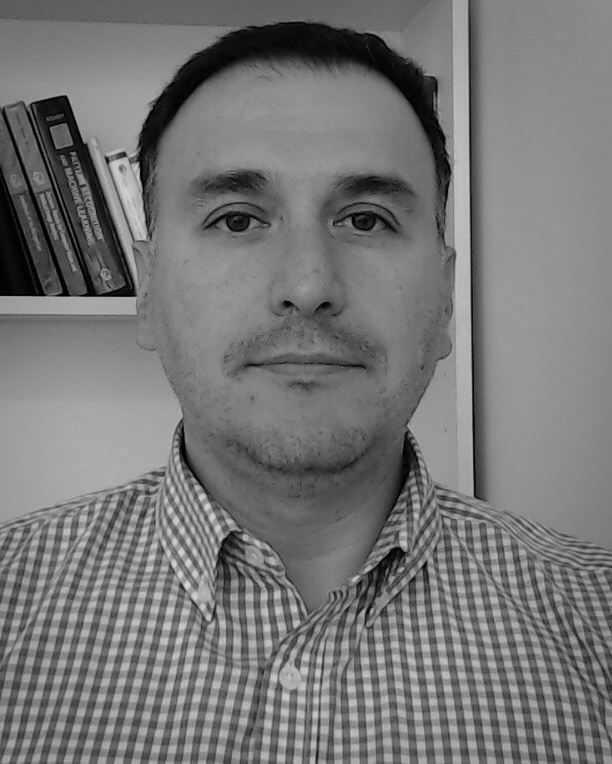}}]{Juan Tapia} received a P.E. degree in Electronics Engineering from Universidad Mayor in 2004, a M.S. in Electrical Engineering from Universidad de Chile in 2012, and a Ph.D. from the Department of Electrical Engineering, Universidad de Chile in 2016. In addition, he spent one year of internship at University of Notre Dame. In 2016, he received the award for best Ph.D. thesis. From 2016 to 2017, he was an Assistant Professor at Universidad Andres Bello. From 2018 to 2020, he was the R\&D Director for the area of Electricity and Electronics at Universidad Tecnologica de Chile - INACAP. He is currently a Senior Researchet at Hoschule Darmstadt(HDA), and R\&D Director of TOC Biometrics. His main research interests include pattern recognition and deep learning applied to iris biometrics, morphing, feature fusion, and feature selection. 
\end{IEEEbiography}

\begin{IEEEbiography}[{\includegraphics[width=1in,height=1.25in,clip,keepaspectratio]{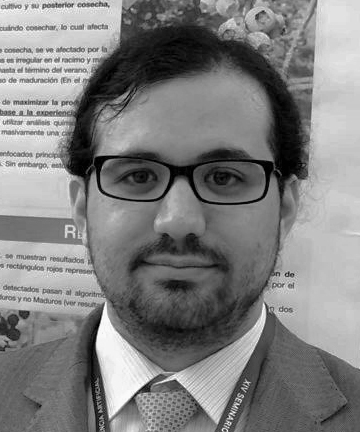}}]{Sebastian Gonzalez} received a B.S. in Computer Engineering from Universidad Andres Bello in 2019. Currently, he is a researcher at TOC Biometrics company. His main interests include computer vision, pattern recognition and deep learning applied to real problems such as tampering detection, classification and segmentation.
\end{IEEEbiography}

\begin{IEEEbiography}[{\includegraphics[width=1in,height=1.25in,clip,keepaspectratio]{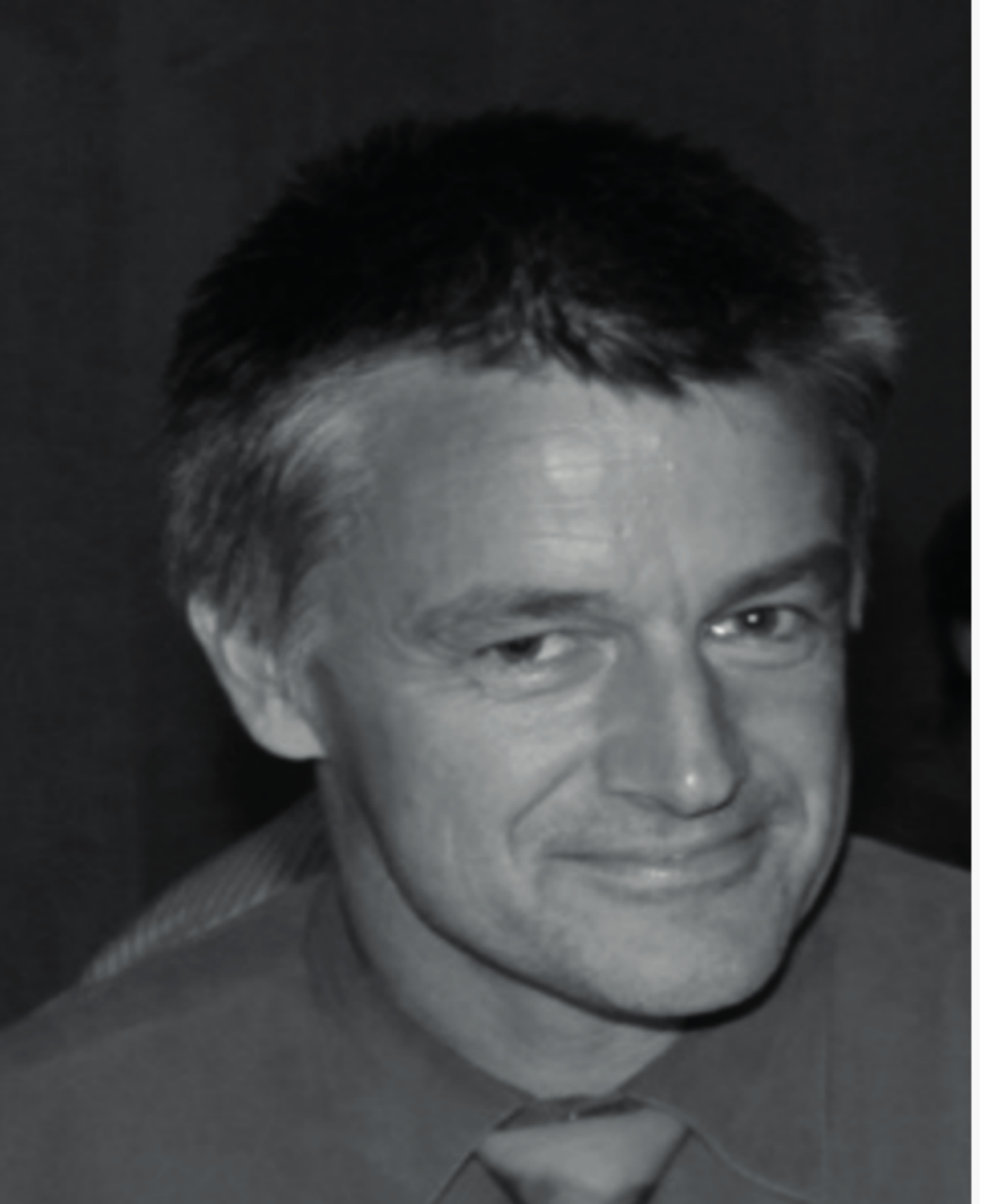}}]{Christoph Busch} is a member of the Department of Information Security and Communication Technology (IIK), Norwegian University of Science and Technology (NTNU), Norway. He holds a joint appointment with the Faculty of Computer Science, Hochschule Darmstadt (HDA), Germany. He coauthored more than 500 technical papers and has been a speaker at international conferences, and served for several conferences, journals, and magazines as a Reviewer such as ACM-SIGGRAPH, ACM-TISSEC, the IEEE COMPUTER GRAPHICS AND APPLICATIONS, the IEEE TRANSACTIONS ON SIGNAL PROCESSING, the IEEE TRANSACTIONS ON INFORMATION FORENSICS AND SECURITY, the IEEE TRANSACTIONS ON PATTERN ANALYSIS AND MACHINE INTELLIGENCE, and the Computers and Security journal (Elsevier). He is also an Appointed Member of the Editorial Board of the IET Biometrics journal and the IEEE TRANSACTIONS ON INFORMATION FORENSICS AND SECURITY journal. 
\end{IEEEbiography}

\end{document}